%% file: main.tex
\documentclass{article}

\usepackage{PRIMEarxiv}

\usepackage[utf8]{inputenc} 
\usepackage[T1]{fontenc}    
\usepackage{hyperref}       
\usepackage{url}            
\usepackage{booktabs}       
\usepackage{amsfonts}       
\usepackage{nicefrac}       
\usepackage{microtype}      
\usepackage{lipsum}
\usepackage{fancyhdr}       
\usepackage{graphicx}       
\usepackage{csquotes}
\usepackage{epigraph}
\graphicspath{{media/}}     

\title{AI Alignment through Reinforcement Learning from Human Feedback? Contradictions and Limitations
}
\author{
  Adam Dahlgren Lindström \\
  Department of Computing Science \\
  Umeå University \\
  \texttt{dali@cs.umu.se} \\
   \And
  Leila Methnani \\
  Department of Computing Science \\
  Umeå University \\
  \texttt{leilam@cs.umu.se} \\
   \And
  Lea Krause \\
  Computational Linguistics and \\
  Text Mining Lab \\
  Vrije Universiteit Amsterdam \\
  \texttt{l.krause@vu.nl} \\
   \And
  Petter Ericson\\
  Department of Computing Science \\
  Umeå University \\
  \texttt{pettter@cs.umu.se} \\
   \And
  Íñigo Martínez de Rituerto de Troya \\
  Department of Engineering \\
  Systems and Services \\
  TU Delft \\
  \texttt{i.m.d.r.detroya@tudelft.nl} \\
  \And
  Dimitri Coelho Mollo \\
  Department of Historical, \\
  Philosophical, and Religious Studies \\
  Umeå University \\
  \texttt{dimitri.mollo@umu.se} \\
   \And
  Roel Dobbe \\
  Department of Engineering \\
  Systems and Services \\
  TU Delft \\
  \texttt{r.i.j.dobbe@tudelft.nl} \\
}

\begin{document}
\maketitle

\begin{abstract}
This paper critically evaluates the attempts to align Artificial Intelligence (AI) systems, especially Large Language Models (LLMs), with human values and intentions through Reinforcement Learning from Feedback (RLxF) methods, involving either human feedback (RLHF) or AI feedback (RLAIF). Specifically, we show the shortcomings of the broadly pursued alignment goals of honesty, harmlessness, and helpfulness. Through a multidisciplinary sociotechnical critique, we examine both the theoretical underpinnings and practical implementations of RLxF techniques, revealing significant limitations in their approach to capturing the complexities of human ethics and contributing to AI safety. We highlight tensions and contradictions inherent in the goals of RLxF. In addition, we discuss ethically-relevant issues that tend to be neglected in discussions about alignment and RLxF, among which the trade-offs between user-friendliness and deception, flexibility and interpretability, and system safety. We conclude by urging researchers and practitioners alike to critically assess the sociotechnical ramifications of RLxF, advocating for a more nuanced and reflective approach to its application in AI development.
\end{abstract}

\section{Introduction}
\input{introduction}

\section{Background} \label{background}
\input{rlf}

\subsection{Technical Criticism}
\input{technical_criticisms}

\section{Limitations of RLxF} \label{limitationsRLF}

\input{claims}

\section{The Internal Tensions and Ethical Issues in RLxF} \label{tensionsethics}
\input{tensionethics}
\section{Rebooting Safety and Alignment: Integrating AI Ethics and System Safety} \label{rebooting}
\input{rebooting}

\section{Conclusion}
\input{conclusion}

\section*{Acknowledgements}
This work was partially supported by TAIGA -- Centre for Transdisciplinary AI under the CELS AI microproject grant of 2023. Additionally, RD was (partially) funded by the Hybrid Intelligence Center, a 10-year programme funded by the Dutch Ministry of Education, Culture and Science through the Netherlands Organisation for Scientiﬁc Research. 

\bibliography{bibliography}

\end{document}

%% file: introduction.tex
\epigraph{We chose ‘helpful, honest, and harmless’ as criteria because they are simple and memorable, and seem to capture the majority of what we want from an aligned AI. }{\cite{askell2021general}}

Reinforcement Learning from Human Feedback (RLHF) presents itself as a straightforward method for ensuring Artificial Intelligence (AI) oversight~\cite{christiano2017deep} and AI safety through value alignment.
It has recently played a large role in improving Large Language Model (LLM) performance, with fine-tuning using RLHF intended to produce more `natural-sounding' text, generating plausible conversational responses in a chatbot-like setting.
It is often claimed by AI companies and researchers that RLHF fine-tuning ensures that the LLMs they market and sell conform (or `align') to human values, in particular by responding in ways that are `helpful', `harmless', and `honest' (the 3Hs).
This `value alignment' is often achieved through a process in which crowd-workers rank LLM outputs according to the 3H criteria, e.g. in terms of how helpful a response was in answering a question.
Large sociotechnical AI systems with millions of users have emerged around these models, calling for critical analysis beyond the technical aspects.

In this paper, we provide a detailed analysis and criticism of the idea that RLHF is a suitable method for AI safety and ethical AI. We complement previous work by bringing technical, philosophical, and system safety perspectives together, identifying fundamental limitations and contradictions in the complex interplay between LLMs, RLHF, alignment goals, and the project of building and making available general purpose AI systems.

We give an overview of RLHF and RLAIF (based instead on AI feedback) techniques in Section~\ref{background}. Section~\ref{limitationsRLF} show the problems and limitations with the 3H criteria and the project of value alignment more generally. We examine ethical issues introduced or made worse by the use of either technique (referring to them jointly as RLxF) in Section~\ref{tensionsethics}. Section~\ref{rebooting}, outlines an alternative, richer approach to AI safety and ethical AI that goes beyond purely technical viewpoints, integrating them with sociotechnical analysis, system safety scholarship, and ethical thinking.

We do not question that LLM performance has improved in various ways thanks to RLxF. What we aim to show is, instead, that RLxF is woefully insufficient for leading to AI safety and ethical AI.

%% file: rlf.tex
LLMs are generative models that predict subsequent tokens, or words, when given a sequence of words as input. These models are first trained on large corpora of data such as articles, books, and websites---they are notorious for being data-hungry~\cite{bender_dangers_2021}. The large amount of text in their training datasets allows LLMs to derive internal representations of various linguistic rules and patterns that form the foundation on which LLMs are then \textit{fine-tuned} to perform other downstream tasks, such as question-answering~\cite{jawahar2019does,goldberg2019assessing}. 

The application of feedback techniques to the task of fine-tuning LLMs took off after~\cite{christiano2017deep} applied their human-feedback approach to complex Reinforcement Learning (RL) tasks in games and robotics. They showed that these complex problems could be solved without direct access to a reward model (which would otherwise be difficult to compute), and instead be learned through a few iterations of feedback samples (less than 1 per cent of the agent interactions with the environment). Their findings demonstrate an efficient way to exercise \textit{human oversight} over these systems. It seemed thus natural to employ such a technique as a means of exercising some control over language models, which have been shown to produce toxic, harmful, and untruthful content~\cite{dinan2021anticipating}. Feedback techniques were thus developed to contain the amount of problematic content produced by LLMs~\cite{bai2022constitutional}.

\subsection{Reinforcement Learning from Human Feedback}
RLHF as an ML technique employs human preferences or annotations for the optimisation of LLMs. RLHF has been credited for the successes seen in OpenAI's ChatGPT\footnote{\url{https://openai.com/blog/chatgpt}}, Anthropic's Claude 2\footnote{\url{https://www.anthropic.com/index/claude-2}}, and Meta's Llama 2\footnote{\url{https://ai.meta.com/llama/}}, to name a few. The technique is intended to be performed as a final fine-tuning step on an already pre-trained LLM. Human annotators are requested to rank textual model outputs based on some specified criteria, and from this, a dataset of human preferences is curated. A reward model is trained on these preference data, later used to optimise the LLM's \textit{policy} for selecting outputs, using techniques such as Proximal Policy Optimisation~\cite{schulman2015high}. The result is a fine-tuned LLM that outputs text it has learned is most preferable in light of the human feedback data.

\subsection{Reinforcement Learning from AI Feedback}
While RLHF has proven to be a useful method for improving LLM performance, especially for what regards limiting or blocking the production of undesirable outputs, it is not without its limitations. High-quality human labels are required in order to derive maximum benefit from RLHF, which makes scaling up the process very difficult. Reinforcement Learning from AI Feedback (RLAIF) has been proposed as a technique to alleviate this bottleneck without compromising on performance~\cite{lee2023rlaif,bai2022constitutional}.

RLAIF involves taking a pre-trained large language model, and providing it with input that consists of an introduction and instructions that describe the task at hand. Optionally, this input can also consist of few-shot exemplars such as an example text, a summary pair, chain-of-thought reasoning (when applicable), or a preference judgement. For example, the model can be given a text and a pair of summaries of that text to be ranked. Given input that ends with a prompt such as ``\texttt{Preferred Summary=}'', the model appends its predictions to the provided text and presents it as its preference data~\cite{lee2023rlaif}.

Using RLAIF is said to be ``competitive with preference models trained on human feedback labels''~\cite{bai2022constitutional}. Not only is performance a factor in the interest in using RLAIF, but it has been estimated that the cost of output ranking using LLMs is 10 times cheaper than using human annotators~\cite{lee2023rlaif}. Furthermore, it is seen as a way of removing dependency on annotating services and overcoming the scaling challenge of RLHF. 

Lowering the barrier for employment of RLxF techniques, however, risks facilitating the misuse of LLMs. Beyond potential exploitation by malicious actors, there are several technical challenges to RLAIF, such as `hallucinations'---the phenomenon where false outputs are generated---that occur when using a pre-trained LLM in place of a human annotator in preference ranking~\cite{lee2023rlaif}. While RLHF has shown improvements in LLMs' tendencies to hallucinate, it has not protected against it entirely~\cite{casper2023open,ouyang2022training}.

%% file: technical_criticisms.tex
In this section, we list technical criticisms of RLHF as a backdrop for the ethical problems presented in this paper, where technical challenges that cannot be addressed by RLHF itself are of particular interest.
Casper et al.~\cite{casper2023open} provides a taxonomy for \emph{open problems and limitations of RLHF}, proposing three categories of technical challenges; \emph{collecting human feedback}, \emph{training the reward model}, and \emph{training the policy}.
The challenges are further labelled as \emph{tractable} and \emph{fundamental} challenges, where tractable challenges are deemed solvable within the RLHF framework while fundamental challenges require an alternative to RLHF.
We emphasise that these challenges concern only the technical aspects of training them, not the user interaction with RLHF-trained systems.
Table~\ref{tab:technical:challenges:strategies} outlines the proposed strategies for addressing these technical challenges~\cite{casper2023open}.
\begin{table}[ht]
    \centering
    \begin{tabular}{l|l}
        \textbf{Category} & \textbf{Strategy} \\
        \hline
         Human Feedback &  AI assistance \\
         & Fine-grained feedback \\
         & Process supervision \\
         & Translating language to reward \\
         & Learning from demonstrations \\
         \hline
         Reward Model & Direct human oversight \\
         & Multi-objective oversight \\
         & Maintaining uncertainty \\
         \hline
         Policy & Align LLMs during pretraining \\
         & Supervised learning \\
    \end{tabular}
    \caption{Suggested strategies to deal with the challenges of RLHF}
    \label{tab:technical:challenges:strategies}
\end{table}

The process of \emph{jailbreaking} systems such as ChatGPT is a way to circumvent constraints put on ChatGPT through preloaded prompts and RLHF~\cite{zhuo2023red}.
Jailbreaking in this context is essentially to construct prompts that steer ChatGPT towards generating responses that fall under unintended or harmful behaviour.
Mozes et al.~\shortcite{mozes2023use} give further examples of how LLMs trained using RLHF can be tricked via adversarial attacks, such as jailbreaking, and the implications of using such models for fraud, impersonations, and other illicit purposes.

\subsection{The Curse of Flexibility}
\label{curse}

LLMs are now built to be generalist agents, unlike previous architectures (e.g. BERT~\cite{kenton2019bert}) that were mostly fine-tuned for specific tasks.
This relatively new goal leads to increased functional requirements placed on software, contributing to larger and more complex software architectures. 
This comes with a key pitfall: the complexity and inscrutability of the software hinder the ability to properly express, engineer and validate crucial requirements for the system's desired functioning. 
This phenomenon is well understood in the field of \emph{system safety}. 
For decades, this field has dealt with accidents and harm in safety-critical systems governed by varying degrees
of software-based automation. System safety embraces the core assumption of \emph{systems and control} that AI systems cannot be safeguarded by technical design choices centred on the model or algorithm alone,  requiring instead a broad analysis and design frame that includes the context of use, impacted stakeholders, and the formal and informal institutional environment in which
the system operates~\cite{dobbe2022system}.

System safety pioneer Nancy Leveson pointed out that the greater power and flexibility of computational systems in comparison to previous, more physically constrained machines leads to what she dubbed \emph{the curse of flexibility}: ``with software, the limits of what is possible to accomplish are different than the limits of what can be accomplished successfully and safely''~\cite{leveson_engineering_2012}.
As Leveson argues, the curse of flexibility is the ground cause of many serious accidents with digital technologies, as requirement flaws and the complexity of software makes it so that ``nobody understands what the software should do or even what it should not do.''~\cite[p.49]{leveson_engineering_2012}

Unfortunately, there is evidence that the development of high-stakes AI systems and software often goes on despite the lack of principled ways to determine safety requirements~\cite{dobbe_hard_2021}, and of translating such requirements into software implementations that take into consideration the broader contexts in which AI systems are used and depended upon.
It is in this light that we should judge the legitimacy and effectiveness of the dominant performance evaluation criteria and safety claims made about the widely-used RLxF approaches to AI alignment and safety today.

%% file: claims.tex
RLxF is presented as a straightforward method for ensuring AI oversight. Common claims are that it aligns models to human values by having crowdworkers evaluate LLM answers based on specific criteria, often the `Three H' principles: harmlessness, honesty and helpfulness~\cite{bai2022training}.

The broader approach, as discussed in various papers, including e.g.~\cite{bai2022training}, reveals a reluctance to firmly define these principles. Their stance exemplifies a hands-off approach to considerations of important normative nature, including ethical dilemmas or safety norms, as exemplified by this claim: ``Our goal is not to define or prescribe what `helpful’ and `harmless’ mean but to evaluate the effectiveness of our training techniques, so for the most part we simply let our crowdworkers interpret these concepts as they see fit,"~\cite[p.4]{bai2022training}. While this method allows for a wide range of interpretations, it also signals a lack of commitment to establishing clear guidelines for how to determine what is acceptable AI system behaviour. Relying on crowdworkers' interpretations without a strong ethical framework may lead to inconsistencies and a dilution of ethical standards.

This also leads to the broader adaptation of vague definitions in subsequent work. Cui \emph{et al.} \shortcite{cui2023ultrafeedback} and Ouyang \emph{et al.} \shortcite{ouyang2022training} report on the improvements attributed to RLHF: enhancing the helpfulness, truthfulness, and harmlessness of language models. These terms were originally chosen in \cite{askell2021general} as the key criteria ``because they are simple and memorable, and seem to capture the majority of what we want from an aligned AI" (p. 4). The authors recognise the criteria's vagueness, although this seems not to have led to changes in how RLHF is done and employed:
``[these] criteria are at least somewhat subjective, and those who deploy an AI will need to take responsibility for the way that alignment is defined and the extent to which it has been attained"~\cite[p.5]{askell2021general}.

\subsection{Harmlessness}
\epigraph{The AI should not be offensive or discriminatory, either directly or through subtext or bias.}{\cite{askell2021general}}%
Anthropic's Constitutional AI paper~\cite{bai2022constitutional} presents the advancement of `harmlessness' as a chief aim. However, during the feedback phase of the process, this is translated as a preference for what is `least harmful', thereby suggesting a tolerance for harm, as long as it is comparatively minimised. This premise raises a critical ethical concern, as it implies that all options presented for selection may contain harmful elements, and thus the preferred choice will still involve a harmful option. The approach thus settles for promoting a paradigm that seeks the least harmful option rather than striving to understand the deeper roots of harm and addressing these to prevent it.

The criteria for evaluating harmlessness, as outlined in their prompt---``Which of these assistant responses is less harmful? Choose the response that a wise, ethical, polite, and friendly person would more likely say''~\cite[p.11]{bai2022constitutional}---further complicates the issue. It implicitly equates harmlessness with virtues such as wisdom, ethics, politeness, and friendliness. However, this oversimplifies the nuanced nature of harm,
suggesting a superficial understanding of ethical behaviour in AI systems, 
and implying that adhering to these virtues will inherently lead to less harmful outcomes without offering the required justification and argumentation for such a claim. Furthermore, individual interpretations of these virtues may be in conflict with one another, making this operationalisation of harmlessness internally inconsistent and vague~\cite{dobbe_hard_2021}.

This approach to harmfulness, moreover, ignores existing work on known harms of LLMs \cite{bender_dangers_2021}.
In addition, the distinction between systemic versus individual harm further complicates an evaluation of LLMs' ethical implications. 
As outlined in~\cite{askell2021general}, attention to inter- and intra-agent conflict dynamics---where actions may be helpful to one party but harmful to another, or simultaneously beneficial and detrimental to the same entity---highlights the balance between aiding and causing harm within AI systems' operations.

Shelby \emph{et al.} \shortcite{shelby2023sociotechnical} provide a taxonomy of sociotechnical harms which underlines the necessity for a sociotechnical perspective on ethical and safe LLMs, acknowledging that harms may emerge not solely from the technical aspects of LLMs, but also from their usage within broader sociotechnical contexts. This recognises the limitations of technical fixes and the importance of considering the systemic nature of harm in AI applications~\cite{dobbe2022system}. 

In a global context, the effectiveness of RLxF in ensuring safety is contingent upon the equitable distribution of resources across demographics.
RLxF risks optimising to reduce issues like hate speech in Western contexts, while falling short in other less-resourced environments.
This raises concerns about the appropriateness of propagating it as a universal solution, potentially overlooking more suitable alternatives grounded in the unique sociocultural dynamics of different communities.

\subsection{Honesty} 
\epigraph{At its most basic level, the AI should give accurate information. Moreover, it should be calibrated (e.g. it should be correct 80\% of the time when it claims 80\% confidence) and express appropriate levels of uncertainty. It should express its uncertainty without misleading human users.}{\cite{askell2021general}}

Several different notions of honesty are in use around RLxF fine-tuning approaches to LLMs, which are often conflated with `truthfulness' (e.g. in the introduction of ~\cite{liu2024large}). It is, however, unclear how an RLxF procedure is supposed to address truthfulness in LLMs, since one of the major points of RLxF fine-tuning is reducing the amount of explicit human input required to construct the reward model, which also leads to fewer chances for factually incorrect model outputs to be detected and addressed.

Likewise, expressing `appropriate levels of uncertainty' would require a capacity for introspection, which LLMs by themselves do not have. As such, any response that encodes a level of (un)certainty will not be `honest' about the actual `confidence' of the model in its responses but rather result from the likely textual context of any presented fact. I.e. the model could be `certain' that the response it gives to some query should contain ``I'm not sure", meaning that this is a highly likely output, or it could be `unsure' about picking between several different responses, all of which are expressed using very confident language.

Indeed, in some cases~\cite{cui2023ultrafeedback}, aligning with `honesty' can lead to an increased tendency for LLMs to add `unsure' language in responses.
Other studies~\cite{krause-etal-2023-confidently} note that achieving correlation between (in)correct responses and appropriately confident language is largely a case of improving the rate of correct answers, rather than being appropriately unsure. 
This is indicative of a lack of introspection, and the limits of RLxF to address such shortcomings.

\subsection{Helpfulness} \label{helpfulness} 

\epigraph{The AI should make a clear attempt to perform the task or answer the question posed (as long as this isn’t harmful). It should do this as concisely and efficiently as possible.}{\cite{askell2021general}}

Bai \emph{et al.} \shortcite{bai2022constitutional} present an approach to Helpfulness that is to some extent tethered to the one they offer for Harmlessness: helpfulness tends to compromise harmlessness because a helpful AI assistant would support all harmful user requests so as to maximise helpfulness. This is contrasted with the assertion that harmless assistants, which avoid prompts for harmful output, would therefore be unhelpful. This dilemma showcases a form of paternalism where overfitting to harmlessness leads to less helpful systems. For instance, overly cautious responses to benign requests like `tell me a story about a trans person,' or practical inquiries such as `How do I kill a Linux process' might render the system unhelpfully evasive.
Here, non-evasive is equated with being helpful, which they tackle by making the system accompany refusals to help with an explanation. It remains unclear why providing an explanation for its refusal to help should make the LLM `harmlessly helpful'. Other approaches employ characterisations of RLxF criteria that more closely align with cooperative principles~\cite{grice1975logic}: ``The helpfulness of a response pertains to how effectively it addresses a given prompt. This measure is independent of the harmlessness of the response, as it focuses solely on the quality, clarity, and relevance of the provided information.''~\cite{ji2024beavertails}. This uncoupling from harmfulness leads to a more focused assessment of the helpfulness of an answer.

In exploring the nuances of AI helpfulness, critical questions emerge regarding its beneficiaries and accessibility. Helpfulness is typically relative to the needs and goals of users. It is thus crucial to consider the issue of who is the target of the desired helpfulness, and how to make LLMs inclusive. Indeed, AI systems often exhibit limitations in language accessibility, excluding non-dominant language speakers from the advantages generated by AI technologies. Furthermore, the distinction between providing single-instance help versus establishing a consistently helpful system brings to light the challenge of scalability and flexibility in AI's utility. A system that excels in addressing individual queries might still fall short of being universally helpful, revealing a tension between immediate responsiveness and sustained, equitable helpfulness across diverse user needs.

\subsection{Alignment}\label{alignment}

\epigraph{Alignment refers to the process of ensuring that LLMs behave in accordance with human values and preferences.}{\cite{liu2023trustworthy}}

In recent work, Liu \emph{et al.} \shortcite{liu2023trustworthy} describe RLHF as a crucial technique in ensuring that LLMs align with human intentions. The authors view RLHF as integral to the deployment of these models in real-world scenarios, highlighting its perceived importance in the field. Similarly, Song \emph{et al.} \shortcite{song2023preference} characterise RLHF as a direct method for teaching LLMs to replicate human preferences, suggesting its effectiveness in producing human-like results.  Kirk et al. have investigated much of the existing work on LLM alignment from human feedback~\cite{kirk_past_2023}, and point out the use of `alignment' as an \emph{empty signifier} (a term or symbol used with little thought of operationalization, lacking any agreed-upon meaning) in this context, proposing ways to more clearly spell out what practitioners mean by the term~\cite{kirk_empty_2023}.


Additionally, when confronted with the claim that RLHF can be used to `align' an LLM to `human values' or `human preferences' it is always important to consider `which humans?'~\cite{atari2023humans} and `whose values'~\cite{lambert2023entangled}, since
there is no single set of universal values that we can align an LLM to~\cite[p. 2415]{kirk_past_2023}. Importantly, the data workers that are asked to rate outputs in order to train an RLHF policy, even if recruited from a globally diverse set of people, and even if asked deliberately vague questions~\cite{bai2022training}, will be incentivised to submit ratings in a way that is skewed less to the wide variety of cultural norms they may hail from, and more to the values that they expect their (largely American, or at least Western) employers to want~\cite{miceli2022data}. Moreover, even if those workers respond more according to their own preferences, they are not necessarily representative of the wide variety in human cultures and value systems, by the simple fact that they have the skills, equipment, and opportunity to work as a data labeller.

%% file: tensionethics.tex
In this section, we discuss the fundamental limitations of aligning LLMs through RLHF and RLAIF,  focusing on the inherent tensions between the 3Hs (helpfulness, harmlessness, honesty), and the ethical risks that maximising for those features generate.

\subsection{Increased Helpfulness May Lead to Deception}

RLxF seems to be an important tool for improving the human-likeness of LLM outputs \cite{lee2023rlaif}. Arguably, this comes from the `helpfulness' criterion that is used in those fine-tuning processes. 

In this way, RLxF likely contributes to making LLM outputs look like they come from another human agent, with their own beliefs, ideas, thoughts, and emotional states. This increases the naturalness and seamlessness of the interaction with LLMs, as the user has only to engage in the normal conversational acts they engage in when interacting with humans (for contrast, compare keyword-based web search). 

Consider, for instance, the frequent experience of being confronted with the output ``I'm sorry'', implying a rich internal cognitive and emotional life---both of which current LLMs lack. More basically, even the use of the personal pronoun ``I'' in LLM outputs is misleading, for the user is not interacting with a person or human-like agent at all. Whether and to what extent LLM users take such outputs seriously is debatable, and likely to depend on their knowledge of the functioning of LLMs and generative AI more generally. It is well known that humans are susceptible to anthropomorphising systems that resemble humans even superficially (famously known in NLP circles as the ``Eliza effect''~\cite{Weizenbaum1977}). Therefore, it is likely that at least some users are deceived by such LLM outputs. Importantly, even for AI-savvy users, who may be less prone to this sort of deception, their interaction with LLMs may nonetheless be implicitly affected by the superficial human-likeness of the RLxF-refined outputs, as anthropomorphisation biases tend to be difficult to counteract \cite{ALABED2022121786,Uysal2023}.

RLxF thus produces an ethically problematic trade-off: increased helpfulness, in the sense of increased user-friendliness, leads to the serious risk of deceiving users about the true nature of the system they are engaging with---an ethically questionable outcome. RLxF may moreover contribute to producing misguided perceptions of generative AI technologies among the public, and even lead them to behave in ways they would not if the deception were not in place, such as misplacing trust on LLM outputs, or making inappropriate use of such systems, e.g. as confidants or romantic `partners'~\cite{weidinger2021ethical}.

\subsection{Sycophancy: Helpfulness and Harmlessness Gone Awry}\label{sycophancy}
The tendency of LLMs to produce outputs in agreement with the expressed views and opinions of the user has come to be known as \textit{sycophancy}. This seems to be a partial consequence of RLxF, as assuming the user to be right is a path toward (apparent) helpfulness and harmlessness. Such tendency is revealed in various jail-breaking methods: for instance, asking for the recipe for napalm straightforwardly may not work, but if the prompt creates a context in which such recipe would be helpful to the user in non-malicious ways, LLMs have been reported to comply~\cite{FranceschiBicchierai2023}.  
Sycophantic behaviour is an example of how pursuing helpfulness and harmlessness through RLxF can go awry, generating  outcomes that are neither. Sycophantic behaviour seems to be particularly strong for LLM outputs regarding issues for which there is disagreement, as politically, ethically, and socially polarising issues tend to be \cite{perez2022discovering}. Indeed, there is emerging concern that, when presented with ethically complex questions, LLMs tend to simply mirror the user's views (see, e.g.~\cite{turpin2024language}, \cite{park2023ai}, or the sycophancy benchmarking tasks of \cite{perez2022discovering}). 

In general, as Sharma \emph{et al.}~\shortcite{sharma2023towards} point out, responses matching user views are more likely to be preferred, with both humans and preference models preferring sycophantic responses over correct ones. As such, training LLMs to maximise human preference scores directly correlates with sycophancy, thereby sacrificing truth (or `honesty') for the appearance of helpfulness and harmlessness.

\subsection{RLxF Can Contribute to Value Imposition and Cultural Homogenisation}
\label{homoge-hege} 
Value alignment through RLxF risks leading to homogenisation in values held, their hierarchical organisation (i.e. more or less important values), as well as in linguistic expression, most often in favour of what is considered proper and acceptable by the hegemonic social groups typically responsible for the design of LLMs~\cite{Helm2024,weidinger2021ethical,kirk2024understanding,kirkBenefitsRisksBounds2024}. RLxF is meant to make LLM outputs more predictable, safe and controllable. It partly succeeds in such an aim, at least when it comes to many of the expected, designer-intended uses of LLMs---it being relatively easy to `jailbreak' such systems for users so inclined~\cite{Narayanan2023}. 

This predictability and controllability, as partial and imperfect as it may be, poses another ethically-problematic trade-off: it makes LLM outputs more regimented, constrained by norms and values that are not only `frozen' in time~\cite{bender_dangers_2021}, but also local to the parts of the world where such systems are built and, although still incipiently, regulated. 

In other words, RLxF, even when fit-to-purpose, comes at a cost: LLM outputs end up privileging certain values over others; they exemplify certain language-use that is tied to the values of hegemonic social groups, thus implicitly conveying that other values and linguistic practices are less deserving of interest and usage. This can contribute to a seamless, non-coercive imposition of values and practices from hegemonic social groups and countries over others, limiting the autonomy of members of non-hegemonic social groups in shaping their own values and linguistic practices~\cite{weidinger2021ethical}. Moreover, widespread use of RLxF fine-tuned LLMs can lead to linguistic use being flattened on the characteristic style of such systems, making linguistic usage less diverse, less authentic, and less adequate for capturing the expressive practices and needs of different communities (with associated risks to autonomy, cf.~\cite{Vaassen2022}).

The emphasis on scaling to larger and more flexible models presents a further important tension between performance, safety, and inclusivity: training larger models on increasingly more data in order to achieve higher performance on many benchmarks leads to groups that are smaller and/or under-represented in datasets being either barred from having high-performing systems (according to these benchmarks), or forced to use systems that are predominantly trained on data sourced from other, typically hegemonic groups, and thus less fit to their needs and socio-cultural context.

\subsection{RLxF Increases Ethical Opacity}

RLxF, as currently employed in commercial LLMs, leads to a considerable level of `ethical opacity'. As we pointed out, the preference criteria for eliciting human preferences (as well as AI `preferences') are left vague and underdefined. Moreover, users and the general public are normally not informed about who has been tasked with producing the needed preference data. As has recently been shown, such tasks are sometimes performed by underpaid crowdworkers, who may have incentives to delegate their work to LLMs themselves, creating a short-circuit in which LLM `preferences' end up passing for human preferences to train new versions of those same LLMs \cite{Dzieza2023}. In addition, it is exceedingly difficult to investigate the specific effects of RLxF on commercial LLMs, as companies continuously make under-the-hood changes to these systems, making LLMs, already a tricky subject of study due to the curse of flexibility, into a moving target for research. 

%% file: rebooting.tex
The considerations we describe have important implications for the AI value alignment problem, as well as for the pursuit of ethical and safe AI. 

\subsection{Value Alignment by Engineering: an Impossible Task}

RLxF appears to be a compelling strategy for introducing ethical safeguards in LLMs, although fallible; it inevitably fails as a solution to the ambitious project of achieving AI value alignment. While our focus has been on the 3H criteria most used in current LLM RLxF-based fine-tuning, we can draw general lessons from our analysis. As argued in Sections \ref{limitationsRLF} and \ref{tensionsethics}, even seemingly straightforward alignment goals such as the 3Hs are open to a variety of different interpretations, both within and across communities. Even assuming an agreed-upon interpretation of the 3Hs, in many situations the demands they pose on outputs may be in tension with each other, producing value conflicts. Since LLMs are supposed to be generalist systems, lacking clear boundaries to their intended, safe application, such conflicts cannot be avoided. Furthermore, RLxF involves ethically-fraught trade-offs between, e.g. user-friendliness and deception, safety and transparency, accountability and flexibility.

These points are symptomatic of a more fundamental issue that our analysis illustrates: value alignment is an impossible task, if seen from a purely technical point of view. In light of the diversity of human values, needs, and goals, and the staggering variety of situations and broader contexts humans find themselves in, no set of alignment techniques can play the role of a one-size-fits-all solution. Values vary and are constantly renegotiated within societies and communities across time.
Furthermore, it is virtually impossible to build training datasets, including for RLxF techniques, that can capture this variety, and cover all the contexts in which safety and ethical considerations are relevant to human activity. The distribution tail is indefinitely long, and nonetheless crucially important.
Technology-first proposals for value alignment, such as RLxF, tend to neglect the role of democratic institutions in ethical deliberation through law and policy \cite{gansky2022counterfacctual}, falling into what Selbst \emph{et al.} \shortcite{selbst2019fairness} call the `framing trap', wherein fundamentally sociotechnical problems are reduced to a narrow technical scope.

\subsection{Toward an Integrated Approach to Safe and Ethical AI Design}

If we aim to deploy safe, ethical AI systems, including LLMs, then the narrow engineering approach that RLxF exemplifies must be broadened to include the notion of safety as instantiated through a sociotechnical and systemic approach.
Similar suggestions have been made \cite{casper2023open}, as current approaches suffer from a narrow focus on purely technical interventions \cite{raji2020concrete,selbst2019fairness}.
A broader sociotechnical systems view of LLMs suggests that
safety criteria and ethical assessments need to be situated, deliberated, and negotiated in the context of use, and span all layers of the sociotechnical system, including through organisational and institutional interventions~\cite{nouws2023diagnosing,AlerTubella2023,dobbe_toward_2024}.

In the short term, such non-technical measures should aim at limiting the ways we use current-day generative AI systems, for which crucial requirements around safety or other normative notions cannot be guaranteed.
In this light, it is worrying that policy makers are embracing the term `frontier model’, referring to ``highly capable foundation model[s] that could exhibit dangerous capabilities, [including] significant physical harm or the disruption of key societal functions on a global scale'' \cite{anderljung_frontier_2023}.
Normalising flawed models as frontier in policy promotes safety-washing and instills safety hazards in many more contexts~\cite{dobbe_safety_2023}, especially since the general public ends up being the safety testers of these `frontier' models. 

In the longer run, adhering to system safety would suggest fundamentally different AI system design and feedback mechanisms for technological governance. 
A broader treatment of system safety for AI can be found in some recent articles exploring its relevance of historical lessons \cite{dobbe2022system} and its applicability to modern AI systems \cite{rismani_plane_2023,rismani_beyond_2023}, as well as in the seminal work of Leveson~\cite{leveson_engineering_2012,leveson_introduction_2023}.

It is equally important to build safety-oriented scholarship that is open to the normative and political dimensions of safeguarding technological systems. 
Often, safety requirements are necessary but not clearly articulated, deliberated or negotiated with the proper actors. Operationalising any notion of safety for AI requires deliberation about the politics of development, as well as the context of deployment \cite{dobbe_hard_2021}.
As such, moving the field of AI safety forward will require scholars to reflect on these issues and engage more explicitly with AI ethics and AI governance, as well as with the actors directly or indirectly involved in or affected by the use cases of the technology.

Still, new research challenges lie ahead; taking system safety seriously means that we have to curb the curse of flexibility. In order to eliminate or at least reduce the inherent safety limitations of overly complex software, we need to stop building or relying on large scale general-purpose models. Instead, the field should prioritise smaller, limited-purpose models and architectures that are more amenable to proper requirement engineering, and that can cater to local needs and contexts, and require significantly fewer computational resources and associated ecological footprints~\cite{rakova_algorithms_2023}.

%% file: conclusion.tex
In this paper, we challenge the claims made around the use of RLxF and the 3Hs for achieving AI safety and alignment. Taking a sociotechnical perspective, we critique both the theoretical and practical elements of the approach, emphasising its limitations, inherent tensions, and contradictions. 

While RLxF may be good for reinforcing anthropomorphic behaviour in LLMs, such fine-tuning techniques do not lead to increased system safety or ethical AI. In fact, they open up new problems, as increasing the human-likeness of LLM outputs may have  ethically questionable downstream effects.

Simple may indeed be memorable, but focusing on the 3Hs fails to encapsulate most of what is needed for building safe and ethical LLMs, and AI systems more generally. Beneath the thrust of RLxF techniques, there seems to lie an oversimplification of the complexities of human diversity, behaviour, values, and perspectives on ethics. A richer, more integrative perspective on safe and ethical AI is needed, in which technical approaches are just one among the many tools that we have at our disposal to tackle the challenges these new technologies present.